\documentclass[10pt,twocolumn,letterpaper]{article}

\usepackage{cvpr}
\usepackage{times}
\usepackage{epsfig}
\usepackage{graphicx}
\usepackage{amsmath}
\usepackage{amssymb}
\usepackage{helvet} % DO NOT CHANGE THIS
\usepackage{courier}  % DO NOT CHANGE THIS
\usepackage[hyphens]{url}  % DO NOT CHANGE THIS
\usepackage{bm}
\usepackage[switch]{lineno}
\usepackage{multirow}
\usepackage{xcolor}
\usepackage{colortbl,booktabs}%第二个包定义了几个*rule
\usepackage{graphicx} % DO NOT CHANGE THIS
\usepackage{algorithm}
\usepackage{algorithmic}
\usepackage{epsfig}
\usepackage{amsmath}
\usepackage{amssymb}
\usepackage{bm}
\usepackage{multirow}
\usepackage{graphicx}  % DO NOT CHANGE THIS
\usepackage{caption} % DO NOT CHANGE THIS AND DO NOT ADD ANY OPTIONS TO IT
\usepackage[pagebackref=true,breaklinks=true,letterpaper=true,colorlinks,bookmarks=false]{hyperref}

\cvprfinalcopy % *** Uncomment this line for the final submission

\def\cvprPaperID{1316} % *** Enter the CVPR Paper ID here

\begin{document}

%%%%%%%%% TITLE
\title{Effective and Fast: A Novel Sequential Single Path Search \\ for Mixed-Precision Quantization}

\author{
Qigong Sun, Licheng Jiao, Yan Ren, Xiufang Li, Fanhua Shang, Fang Liu  \\
{\small Key Laboratory of Intelligent Perception and Image Understanding of Ministry of Education, International Research}\\
{\small Center for Intelligent Perception and Computation, Joint International Research Laboratory of Intelligent Perception}\\
{\small and Computation, School of Artificial Intelligence, Xidian University, Xi'an, Shaanxi Province 710071, China} \\
{\tt\small  xd\_qigongsun@163.com, lchjiao@mail.xidian.edu.cn, yanren@stu.xidian.edu.cn, xfl\_xidian@163.com,} \\
{\tt\small  fhshang@xidian.edu.cn, f63liu@163.com}
}
%First Author\\
%% omit the following lines up until the closing ``}''.
%% Additional authors and addresses can be added with ``\and'',
%% just like the second author.
%% To save space, use either the email address or home page, not both
%\and

\maketitle
%\thispagestyle{empty}

%%%%%%%%% ABSTRACT
\begin{abstract}
   Since model quantization helps to reduce the model size and computation latency, it has been successfully applied in many applications of mobile phones,
    embedded devices and smart chips.
    The mixed-precision quantization model can match different quantization  bit-precisions according to the sensitivity of different layers to achieve great performance.
    However, it is a difficult problem to quickly determine the quantization bit-precision of each layer in deep neural networks according to some constraints (e.g., hardware resources, energy consumption, model size and computation latency).
    To address this issue, we propose a novel sequential single path search (SSPS) method for mixed-precision quantization,
    in which the given constraints are introduced into its loss function to guide searching process.
    A single path search cell is used to combine a fully differentiable supernet, which can be optimized by gradient-based algorithms.
    Moreover, we sequentially determine the candidate precisions according to the selection certainties to
    exponentially reduce the search space and speed up the convergence of searching process.
    Experiments show that our method can efficiently search the mixed-precision models for different architectures (e.g., ResNet-20, 18, 34, 50 and MobileNet-V2) and datasets (e.g., CIFAR-10, ImageNet and COCO) under given constraints, and our experimental results verify that SSPS significantly outperforms their uniform counterparts.
    %\emph{\textbf{Code can be available in the supplementary materials.}}
\end{abstract}

%%%%%%%%% BODY TEXT
\vspace{-3 mm}
\section{Introduction}
\vspace{-1 mm}

With the development of deep neural networks (DNNs), DNNs have achieved impressive performance in many applications.
Generous computing resources and footprint memory
are urgently required when deeper networks are used to solve various problems.
Moreover, with the rapid development of chip technologies, especially GPU and TPU, the computational frequency and efficiency have been greatly improved. Most scholars use GPUs as the basic hardware platform for network training due to excellent acceleration capability.
However, for low power consumption platforms (e.g., mobile phone, embedded devices and smart chips), whose resources are limited, it is hard to achieve satisfactory performance for industrial applications.
As one of the typical methods for model compression and acceleration, model quantization methods usually quantize full-precision (32-bit) parameters to low-bits of precision (e.g., 8-bit and 4-bit). Even more, we can extremely constrain weights and activation values to binary \{-1, +1\} \cite{Courbariaux2016Binarized, Rastegari2016XNOR} or ternary \{-1, 0, +1\} \cite{Zhu2016Trained}, which can be computed by bitwise operations.
This logic calculation is more suitable for the implementation of FPGA and other service-oriented computing platforms. As implemented in \cite{Liang2017FP}, they achieved about $705\times$ speed up compared with CPU and are $70\times$ faster than GPU in the peak condition.

\begin{figure}[t]
    \setlength{\abovecaptionskip}{0.1cm}
    \setlength{\belowcaptionskip}{-0.3cm}
	\centering
	\includegraphics[width=3.1in]{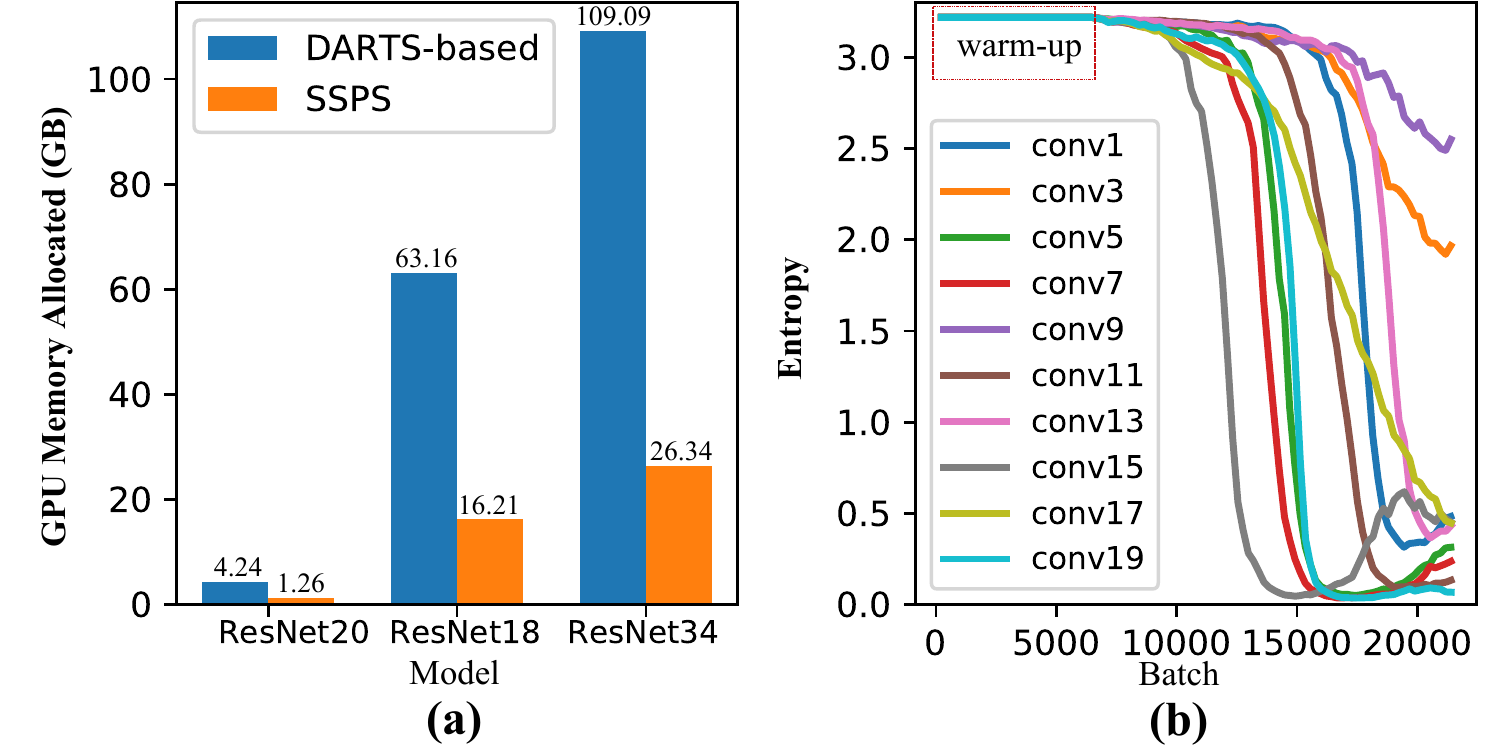}
	\caption{\small{(a) Comparison of the GPU memory usage of searching process on multiple models based on the DARTS-based method and SSPS,
    where each search cell has 5 candidates (i.e., 2-bit to 6-bit) and batch size is 32.
    (b) The searching entropy variation curves of some layers in ResNet-20 on CIFAR-10.}
    }
\end{figure}

In recent years, many methods \cite{choi2018pact,jin2019adabits,li2019additive, lin2017towards,sun2019multi,zhang2018lq} have been proposed to improve the performance of low-precision models.
However, the bit-precisions of most quantization models are set manually based on experience, and all layers have the same quantization precision in general.
Some studies \cite{cai2020zeroq, chen2018searching, dong2019hawq-v2,dong2019hawq} show that different layers have different sensitivity to quantization.
Therefore, the mixed-precision quantization \cite{nandakumar2018mixed,uhlich2019mixed,cai2020rethinking} can achieve better performance
according to the characteristics of network.
Besides, some recent smart chips also support mixed-precision for the DNN inference, e.g., Apple A12 Bionic \cite{2018apple}, Nvidia Turing GPU \cite{2018Nvidia}, BitFusion \cite{sharma2018bit} and  BISMO \cite{umuroglu2018bismo}.

With the successful application of Neural Architecture Search (NAS), \cite{cai2020rethinking,lou2019autoq,wang2019haq} converted the mixed-precision quantization problem as a NAS task and used reinforcement learning or gradient-based methods to search an ideal solution.
Differentiable neural architecture search methods \cite{wu2018mixed,yu2020search} searched over a supernet that contains all candidate architectures
and needs to reside in memory for searching with feature maps.
Fig. 1 (a) shows the GPU memory usage of searching on multiple models based on DARTS \cite{liu2018darts} framework.
Due to its exponential search space with the number of network layers, the searching process requires a lot of hardware resources and is very time-consuming.
In order to relieve the pressure of hardware resources and speed up the searching process,
some methods \cite{wu2018mixed,gong2019mixed} presented a specific small search space, and BP-NAS \cite{yu2020search} was proposed to search
on small datasets and then extend the model to large dataset tasks.
However, the methods rely on the incomplete definition of mixed-precision quantization tasks, which are easy to fall into a local optimal solution.
This task actually seeks a balance between the task-dependent accuracy and given constraints (e.g., energy consumption, hardware resource, quantization precision, model size and bitwise operations).
HAQ \cite{wang2019haq} and DNAS \cite{wu2018mixed} incorporated the complexity cost into loss function and tuned the corresponding balance weight,
which usually takes multiple searches to get an appropriate model.
HAWQ \cite{dong2019hawq} manually chose the bit precision among the reduced search space, and
HAWQ-V2 \cite{dong2019hawq-v2} developed a Pareto frontier based method for selecting the exact bit precision.
Therefore, it is difficult to control the search direction towards the given constraints to meet the deployment requirements.

 In order to address the above issues, we propose a novel differentiable sequential single path search (SSPS) method,
which can quickly find the ideal mixed-precision model of a specific network (e.g., ResNet-20, 18, 34, 50 and MobileNet-V2) satisfying the given constraints (e.g., average weight bit-width and average operation bit-width).
%We introduce the given constraints into loss function to guide the searching process.
%In our method, only one candidate subnet is involved in the calculation at a  training time, thus reducing the dependence of NAS on hardware resources.
The advantages of our method are shown as follows:

%In this paper, we introduce the given expectation into loss function and utilize $Gumbel$-$Softmax$ to solve non-differentiable problem.
%In our experiments, we validate the performance of our method for image classification on CIFAR-10 and large-scale datasets, e.g., ImageNet.

\begin{itemize}
    \item \textbf{Save Resources.}
    We propose a novel differentiable single path search cell, where only one candidate is sampled at a time to carry out calculations.
    That is, it avoids caching all the candidates in memory or participating in calculation together, thus saving hardware resources.
    Fig. 1(a) shows the GPU memory usage of our SSPS method, which is significantly less than that of the DARTS-based method.
    %We use the $Gumbel $- $softmax$ \cite{jang2016categorical} to control the sampling process to achieve
%    the transition from uniform sampling to probabilistic sampling during the searching process.
	\item \textbf{Purpose Search.} We use average weight bit-width and average operation bit-width to measure the given constraints (e.g., model size and bit operations)
    and innovatively introduce them into our constrained loss function.
    By punishing the quantized candidates which deviate from the objective constraint to guide the search direction,
    the problem that the parameters need to be adjusted many times to get a satisfactory solution is greatly alleviated.
	\item \textbf{Fast Search.}
    We use entropy to evaluate the selection certainty of each search cell, and determine the quantization bit-precision of cells sequentially during the searching process.
    %Fig. 1(b) shows the entropy variation curves of some layers in ResNet-20 on CIFAR-10.
%    The convergence speeds of different layers are different.
    Therefore, the complexity is reduced exponentially with the determination of layer quantization bit-precision, and the searching process is significantly accelerated.
    Our method takes less than 7 hours on 4 V100 GPUs to complete a search for ResNet18 on ImageNet, which is faster than DNAS (40 GPU-hours).
    \item \textbf{State-of-the-art Results.} With our proposed techniques applied on a bunch of models (e.g., ResNet-20, 18, 34, 50 and MobileNet-V2) and tasks (e.g., classification and detection), the mixed-precision quantization models we searched are obviously better than other counterparts under the similar constraints.
\end{itemize}

\vspace{-4 mm}
\section{Related Works}

\vspace{-1 mm}
\subsection{Model Quantization}
\vspace{-1 mm}

Model quantization refers to a way to compress and accelerate the model by replacing the full-precision weights or activation values
with fixed-precision values in DNNs.
\cite{Courbariaux2016Binarized,Rastegari2016XNOR} used bitwise operations (e.g., xnor and bitcount) to
effectively compute the matrix multiplication and achieved outstanding efficiency and performance.
%However, the weak expression ability of the binary/ternary value leads to the decline of accuracy in many applications.
To further improve the representation capability,
\cite{sun2019multi, zhou2016dorefa} used the multi-bit quantization to approximate the full-precision weights and activation values.
Most quantizers of multiple bit-widths can be categorized in three modalities: $uniform$ quantizer \cite{choi2018pact,lin2017towards,wang2019haq}, $logarithmic$ quantizer \cite{miyashita2016convolutional} and $adaptive$ quantizer \cite{jin2019adabits,li2019additive}.
According to the quantization granularity of DNNs, model quantization can be divided into $network$-$wise$ quantization \cite{sun2019multi,jacob2018quantization}, $layer$-$wise$ quantization \cite{yazdanbakhsh2018releq,wang2019haq} and $kernel$-$wise$ quantization \cite{lou2019autoq,zeng2019kcnn}.
%The latter two quantization categories can also be regarded as mixed-precision quantization.

The mixed-precision quantization \cite{cai2020rethinking,nandakumar2018mixed,uhlich2019mixed} can match the sensitivity of each layer in DNNs
with  appropriate combination of quantization bit-widths, and it can achieve better results under the same constraints.
HAQ \cite{wang2019haq} added the feedback of acceleration information evaluated by hardware simulator to the training cycle,
and used reinforcement learning to determine the quantization strategy automatically.
\cite{gong2019mixed,wu2018mixed} converted the quantization task to a NAS problem,
and optimized the network weights and architecture parameters by using the back-propagation methods.
However, its pipeline behavior is like DARTS \cite{liu2018darts} at the beginning and it also requires high configuration hardware resources.
\cite{wu2018mixed} spent 40 GPU (V100) hours to complete the search of ResNet-18 in a specific small search space.
By generating distilled data, ZeroQ \cite{lou2019autoq} can fine-tune the models with arbitrary quantization precisions without using any training or validation datasets.
In \cite{yu2019any,guerra2020switchable},
the trained model can match a variety of quantization precisions without any fine-tune or calibration, which will lead to some performance loss.
%BP-NAS \cite{yu2020search} was similar to our method, and introduced the constraints to the loss function.
%However, it is a DARTS-based method, which has a great demand for hardware resources.
%For ImageNet, BP-NAS needs to randomly sample 10 categories with 5000 images as the training dataset to search.

\vspace{-1 mm}
\subsection{Neural Architecture Search}
\vspace{-2 mm}

The emergence of Neural Architecture Search (NAS) breaks the bottleneck of designing neural architectures manually and achieves better performance than human-invented architectures on many tasks, such as image classification \cite{zoph2018learning,liu2017hierarchical}, object detection \cite{zoph2018learning}, semantic segmentation \cite{chen2018searching}, and language models \cite{zoph2017neural,pham2018efficient,dong2019searching}.
The success of NAS requires a variety of search spaces and huge amounts of computing resources,
which makes the optimization of network become a difficult problem.
Commonly used optimization methods are mainly divided into three types: such as reinforcement learning \cite{baker2016designing,zoph2018learning,zhong2018practical}, evolutionary algorithms \cite{real2017large,real2019regularized}, and gradient-based methods \cite{liu2018darts,xie2018snas,ahmed2018maskconnect}.
Besides searching computation operators, NAS methods also search for the width and spatial resolution of each block in the network structure \cite{fang2019densely}.
It can also be used to channel pruning \cite{he2018amc} or filter numbers search \cite{tan2019mnasnet}.
In \cite{cai2018proxylessnas}, the network delay and sparsity are incorporated into the index of search consideration, and it can search the architectures on different tasks (e.g., CIFAR-10 and ImageNet) and different hardware platforms (e.g., GPU, CPU and mobile phones).

\vspace{-2 mm}
\section{Method}
\vspace{-2 mm}

In this paper, we model the mixed-precision quantization task as a NAS problem.
The goal of this task is to find an ideal mixed-precision quantization model for a specific network under some given constraints to meet real-world requirements.
Specifically, the learning procedure of architectural
parameters is formulated as the following bi-level optimization problem:
\setlength{\abovedisplayskip}{5pt}
\setlength{\belowdisplayskip}{5pt}
\begin{eqnarray}
\min_{\bm{\alpha} \in \mathbf{A}, \bm{\beta}\in \mathbf{B}}\mathcal{L}_{val}(\bm{\alpha}, \bm{\beta}, \mathbf{W^*})+ \lambda \mathcal{L_J}_{val}(\bm{\alpha},\bm{\beta}, \mathbf{c}), \\
\!s.t. \quad \mathbf{W^*}(\bm{\alpha}, \bm{\beta})=\textrm{argmin} \mathcal{L}_{train}(\bm{\alpha}, \bm{\beta}, \mathbf{W}) ,
\end{eqnarray}
where $\bm{\alpha}$ and $\bm{\beta}$ represent the architecture parameters for searching of activation values and weights, $\mathbf{A}$ and $\mathbf{B}$ denote the architecture space.
$\mathbf{W}$ and $\mathbf{W^*}$ denote the supernet parameters and selected model weights, $\mathcal{L}_{val}(\,)$ and $\mathcal{L}_{train}(\,)$ represent the task-dependent losses (e.g., the cross-entropy loss) on validation and training datasets, respectively.
$\mathcal{L_J}_{val}(\,)$ measures the constraint loss of the quantization network initialized by $\bm{\alpha}$, $\bm{\beta}$ and $\mathbf{c}$.
$\lambda$ is a super-parameter, and $\mathbf{c}$ is the target vector of given constraints (e.g.,  average weight bit-width and average operation bit-width).
Fig. 2 shows the framework of our proposed SSPS method.

%In order to search the ideal mixed-precision quantization model quickly and purposefully, we propose a SSPS method
%and innovatively introduce some given constraints into its loss function to guide the searching process.
In order to search the ideal mixed-precision quantization model effective and fast, we propose a SSPS method
and innovatively introduce some given constraints into its loss function to guide the searching process.
In this section, we first describe the search space, and then we propose a differentiable single path search cell that constructs one fully differentiable search supernet.
Finally, we describe how we use average weight bit-width and average operation bit-width to evaluate the given constraints and introduce them into our loss function to guide the searching process.
The searching process will be described in the next section.

\begin{figure*}[!htb]
    \setlength{\abovecaptionskip}{0.1cm}
    \setlength{\belowcaptionskip}{-0.5cm}
	\centering
	\includegraphics[width=5.8in]{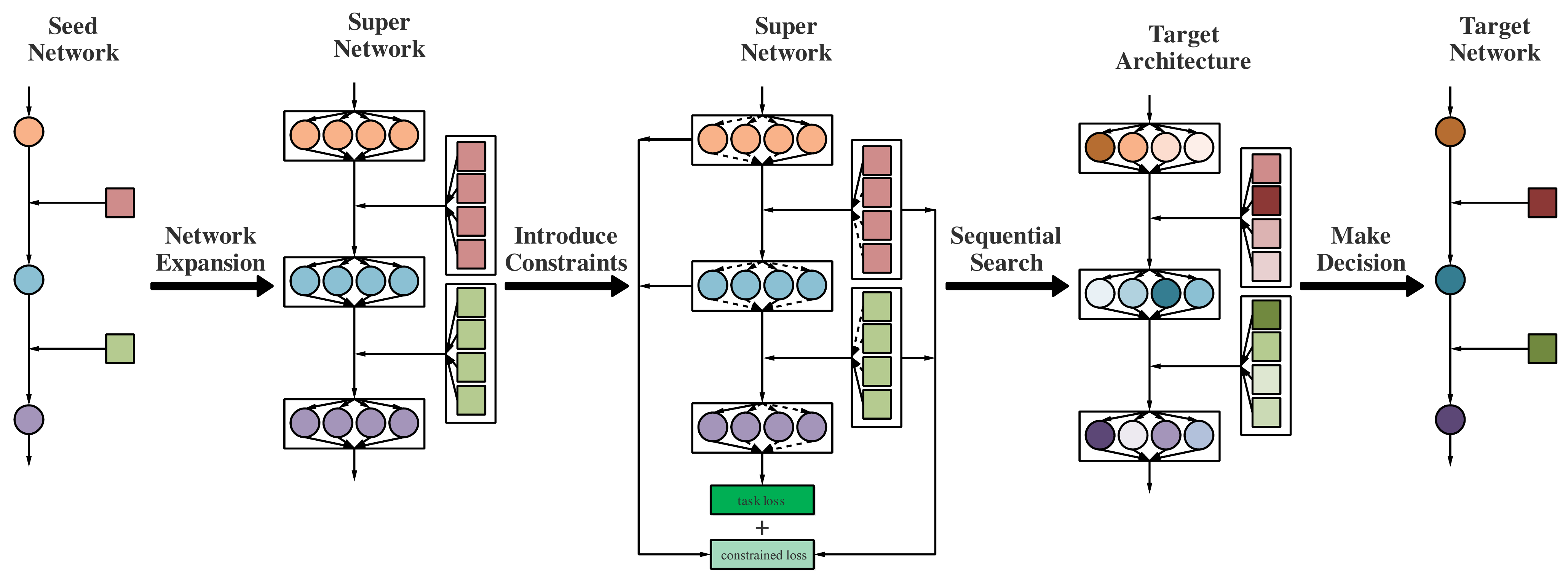}
	\caption{\small{The framework of our proposed SSPS. Firstly, we set a network as the seed network which will be expanded to a supernet.
    In order to search the ideal mixed-precision quantization model, we introduce the given constraints into the loss function
    and use our sequential search method to update the architecture parameters and weights.
    Finally, we select the target network according to the probability of architecture parameters. Best view in color.}
    }
\end{figure*}

\vspace{-1 mm}
\subsection{Weight and Activation Search Cells}
\vspace{-2 mm}

Many recently proposed NAS methods \cite{liu2018darts,cai2018proxylessnas,li2020sgas} focus on cell search (i.e., normal cell and reduction cell).
Once cell architectures are confirmed, %they will be applied in all the layers across the network.
they will stack many copies of these discovered cells to make up a deep neural network.
The purpose of this task is to find the $layer$-$wise$ quantization for specific networks (e.g., ResNet and MobilNet), in which different layers have different quantization precisions.
From Fig. 2, the layer operation contains two search cells (i.e., weight search cell and activation value search cell).

Suppose we use $\mathbf{X}^i$ and $\mathbf{X}^{i+1}$ to represent the input data and output data of the $i$-th layer.
$\mathbf{W}^{i}$ denotes the weights, and $\mathcal{F}(\,)$ denotes the calculation operations (e.g., full connection or convolution).
The computation of those two variables can be formulated as follows:
\setlength{\abovedisplayskip}{5pt}
\setlength{\belowdisplayskip}{5pt}
\begin{eqnarray}
\mathbf{X}^{i+1}=\mathcal{F}(\mathbf{\hat{X}}^i_{m},\mathbf{\hat{W}}^{i}_{n})=\mathcal{F}(\mathcal{BS}_1(\mathbf{X}^i),\mathcal{BS}_2(\mathbf{W}^{i})),
\end{eqnarray}
where $\mathbf{\hat{X}}^i_{m}$ denotes the selected $m$-bit quantized values of $\mathbf{X}^i$ by the bit search cell $\mathcal{BS}_1(\,)$,
and $\mathbf{\hat{W}}^{i}_{n}$ denotes the selected  $n$-bit quantized values of $\mathbf{W}^{i}$ by the bit search cell $\mathcal{BS}_2(\,)$.
%Activation value search cell is shown in Fig. 3.
The general search space $\mathbf{s}^i$ for the $i$-th search cell is shown as follows:
\setlength{\abovedisplayskip}{5pt}
\setlength{\belowdisplayskip}{5pt}
\begin{eqnarray}
m,n \in \mathbf{s}^i = \{1,2,3,4,5,6,7,8,16,32\},
\end{eqnarray}
where all the integers represent the bit-precision. $16$ denotes half-precision floating-point format and
$32$ denotes the single-precision floating-point format.
Thus, the search space size of the whole layer is $100$.
Obviously, the search space of this task $\mathbf{S}=\{\mathbf{s}^1, \mathbf{s}^2, \cdots\}$ is exponential in the number of the model layers $L$,
which is expressed as $10^{2L}$.

\vspace{-1 mm}
\subsection{Differentiable Single Path Search Cell}
\vspace{-2 mm}

Because of the huge search space, reinforcement learning techniques or evolutionary algorithms are computationally expensive and much time-consuming.
DARTS-based methods need to reside all candidate architectures and feature maps in memory.
Thus, they require multiple GPUs with high memory configuration and a small batch size to search.
Fig. 1 (a) shows the GPU memory usage of the DARTS-based method for searching on multiple models.

\begin{figure}[!htb]
    \setlength{\abovecaptionskip}{0.1cm}
    \setlength{\belowcaptionskip}{-0.6cm}
	\centering
	\includegraphics[width=1.8in]{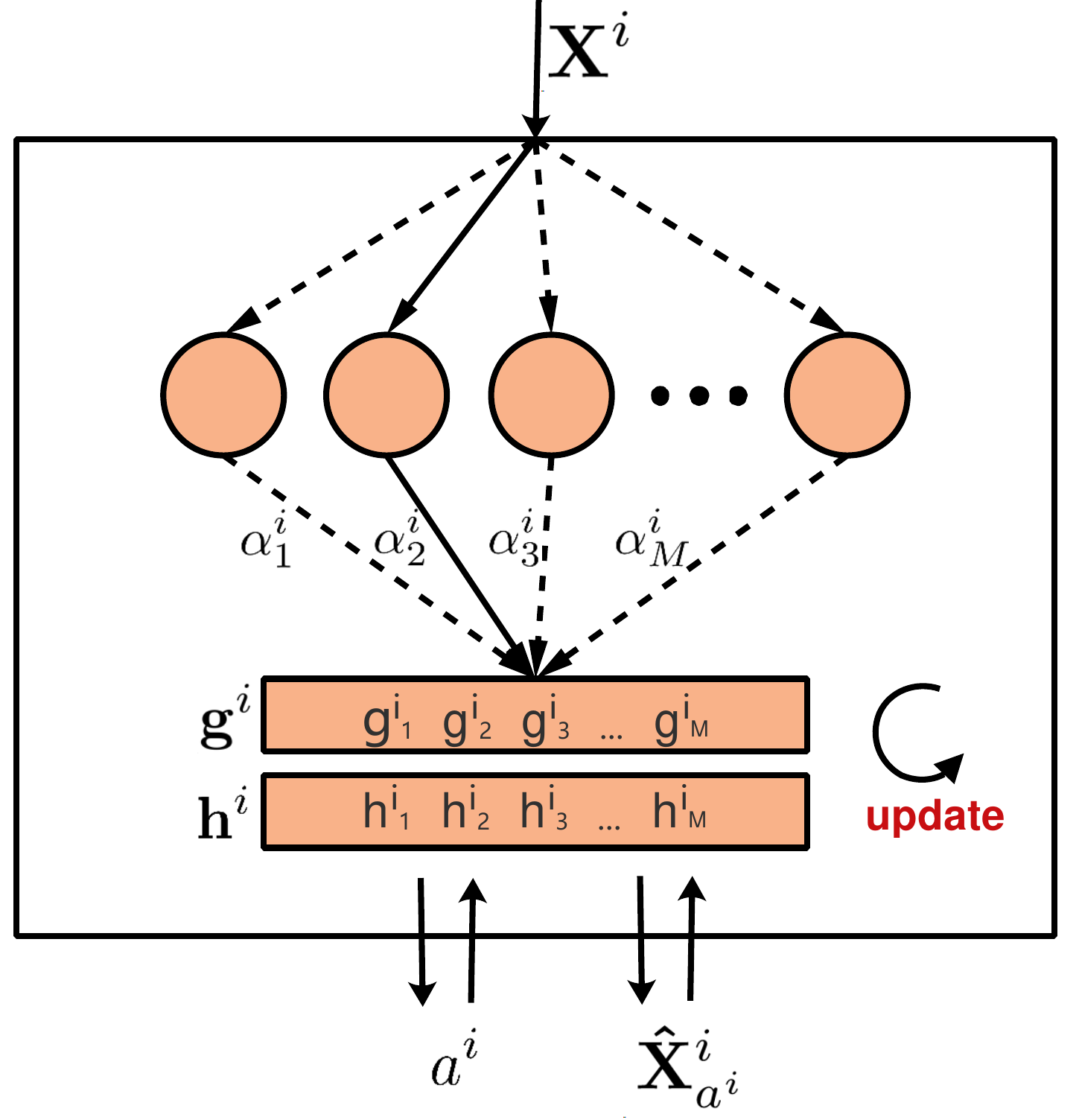}
	\caption{\small{The differentiable single pass search cell for the quantization of activation values,
    where the circles represent the quantization of the different precisions of the input $\mathbf{X}^{i}$,
    and the output $\mathbf{\hat{X}}^{i}_{a^i}$ and $a^i$ denote the quantized values and selected bit-width.}
    }
\end{figure}

In order to save hardware resources and speed up the searching process, we propose a differentiable single path search cell to compose the supernet.
We take activation value quantization as an example, and the search cell is shown in Fig. 3.
The input $\mathbf{X}^{i}$ denotes the full-precision activation values, the output $\mathbf{\hat{X}}^{i}_{a^i}$ and $a^i$ denote the quantized values and the selected bit-width.
We introduce the $Gumbel$-$Softmax$ \cite{jang2016categorical,maddison2016concrete} to control the search strategy.
It approximates the multi distributed sampling process by re-parameterization,
which provides an efficient way to draw samples from a discrete probability distribution.
By this approximation, we can transform the non-differentiable sampling problem into differentiable computation.
Here, we use $\mathbf{g}^{i}$ to represent the sampling probability vector and $g_{m}^{i}$ is the $m$-th element of $\mathbf{g}^{i}$,
which is formulated as follows:
\setlength{\abovedisplayskip}{3pt}
\setlength{\belowdisplayskip}{3pt}
\begin{eqnarray}
g_{m}^{i}=\frac{\textrm{exp}((\alpha _{m}^{i}+o^{i}_m)/\tau)}{\sum_{k=1}^{M}\textrm{exp}((\alpha _{k}^{i}+o^{i}_k)/\tau)},
\end{eqnarray}
where $\alpha _{m}^{i}$ is the $m$-th element of a $M$-dimensional learnable architecture parameter vector $\bm{\alpha}^{i}$,
$o^{i}_k$ is a random variable drawn from the Gumbel distribution ($o^i_k=-\textrm{log}(-\textrm{log}(u))$ with $u \sim \textrm{unif}[0,1]$).
$\tau$ is the temperature coefficient used to control the smoothness of sampling.
Therefore, the activation value search cell can be expressed as follows:
\setlength{\abovedisplayskip}{3pt}
\setlength{\belowdisplayskip}{3pt}
\begin{eqnarray}
\mathbf{h}^{i}&=&\textrm{one\_hot}(\textrm{argmax}(\mathbf{g}^{i})), \\
a^i &=& \sum_{m=1}^{M} \mathbf{s}^i_m h_{m}^{i}, \\
\mathbf{\hat{X}}^{i}_{a^i}&=&\sum_{m=1}^{M}h_{m}^{i}\mathcal{Q}_m(\mathbf{{X}}^{i}) ,
\end{eqnarray}
where $h_{m}^{i}$ is the $m$-th element of one-hot vector $\mathbf{h}^{i}$.
$\mathbf{s}^i_m$ denotes the $m$-th element of the search space.
$\mathcal{Q}_m(\,)$ denotes the $m$-bit quantize function, (e.g., the $uniform$ quantizer).
From Eq. (7), we can get the selected bit precision, which will be used as the input for the constrained loss.
In general, $\textrm{argmax}$ function is used to select the most probable index.
However, since our goal is to sample from a discrete probability distribution, we cannot back-propagate gradients through the $\textrm{argmax}$ function to optimize $\bm{\alpha} ^{i}$.
Here, we use the straight-through estimator (STE) \cite{bengio2013estimating} to back-propagate through Eq. (6).

During the searching process, the real discrete distribution can be approached by gradually reducing the temperature $\tau$.
The higher the temperature, the smoother the distribution.
The lower the temperature is, the closer the generated distribution is to discrete.
At the beginning of the search, it can be regarded as random sampling. With the decrease of $\tau$, it becomes probability sampling.
The resource saving of our search cell can be seen clearly from Fig. 1(a).

\vspace{-1 mm}
\subsection{Constrained Loss Function}
\vspace{-2 mm}

Except for the task-dependent loss (e.g., the cross entropy loss), hardware resources, energy consumption, model size and computational complexity are also important factors affecting real-world applications.
These factors can be effectively controlled by restricting the quantization precision of weights and activation values \cite{sharma2018bit},
and the correlation factors of different hardware platforms are different.
In order to formulate those factors, we introduce average weight bit and average operation bit-width to evaluate model size and bitwise operation,
which are usually applied to evaluate the given constraints \cite{yu2020search,zur2019towards}.
%In order to alleviate the problem that the parameters need to be adjusted many times to get a satisfactory model,
We introduce average weight bit-width and average operation bit-width into our constrained loss function to guide the searching process.

Taking constrain the target model size as expectation, we focus on weight quantization to compress the model.
Generally, model parameters are stored in a 32-bit floating-point type.
When we quantize the weights,  the model size and storage requirements will be reduced.
Suppose we have a model of $L$ layers, and $\textrm{NUM}(i)$ represents the number of parameters in the $i$-th layer.
The average weight bit-width ${E}_{wb}(\bm{\beta})$ can be defined as follows:
\setlength{\abovedisplayskip}{3pt}
\setlength{\belowdisplayskip}{3pt}
\begin{eqnarray}
E_{wb}(\bm{\beta})=\frac{\sum_{i=1}^{L}b^i \textrm{NUM}(i)}{\sum_{i=1}^{L}\textrm{NUM}(i)} ,
\end{eqnarray}
where $\bm{\beta}$ denotes the weight architecture parameter vector, and
$b^i$ is the output of weight search cell that denotes the selected quantization precision of $i$-th layer, the computation of $b^i$ is similar to $a^i$, as shown in Eqs. $(5)\sim(7)$.

The second is to constrain the quantization precision of the weights and activation values to achieve a specific computational complexity.
It is also one of the important reasons affecting industrial applications.
Here, we use average operation bit-width to evaluate bitwise operation computational complexity.
We use $\textrm{FLOP}(i)$ to denote the number of float point operations in the $i$-th layer.
The average operation bit-width ${E}_{ab}(\bm{\alpha},\bm{\beta})$ is related to the architecture parameters $\bm{\alpha}$ and $\bm{\beta}$, and it is formulated as follows:
\setlength{\abovedisplayskip}{3pt}
\setlength{\belowdisplayskip}{3pt}
\begin{eqnarray}
{E}_{ab}(\bm{\alpha},\bm{\beta})=[\frac{\sum_{i=1}^{L}a^i b^i \textrm{FLOP}(i)}{\sum_{i=1}^{L} \textrm{FLOP}(i)}]^{\frac{1}{2}},
\end{eqnarray}
where $\bm{\alpha}$ denotes the architecture parameter of activation value search cell, and $a^i$ is the output of activation value search cell that denotes the selected quantization precision of $i$-th layer.

Based on the above definition, we define a constrained loss function as follows:
\setlength{\abovedisplayskip}{3pt}
\setlength{\belowdisplayskip}{3pt}
\begin{eqnarray}
\mathcal{L_J}(\bm{\alpha},\bm{\beta}, \mathbf{c})=\left \| {E}_{wb}(\bm{\beta}) - c_1 \right \|^2 + \left \| {E}_{ab}(\bm{\alpha},\bm{\beta}) - c_2 \right \|^2,
\end{eqnarray}
where $c_1, c_2 \in \mathbf{c}$ represent the target average weight bit-width and average operation bit-width, respectively.

\vspace{-3 mm}
\section{Searching Process}
\vspace{-2 mm}

%The large search space leads to a high computational complexity of searching process, which causes a long searching period and an incomplete search problem.
Entropy is commonly used to measure the uncertainty of a distribution.
In this paper, we use entropy to evaluate the selection certainty of search cells.
Different entropy values correspond to different selection certainties.
The smaller the entropy, the stronger the selection certainty.
We use the $i$-th activation value search cell as an example, and its probability distribution is computed as follows:
\setlength{\abovedisplayskip}{3pt}
\setlength{\belowdisplayskip}{3pt}
\begin{eqnarray}
\mathcal{P}(\alpha ^i_m)= \frac{\textrm{exp}(\alpha _{m}^{i})}{\sum_{k=1}^{M}\textrm{exp}(\alpha _{k}^{i})},
\end{eqnarray}
where $\bm{\alpha}$ denotes the activation value architecture parameter, and the entropy of this cell is defined as:
\setlength{\abovedisplayskip}{3pt}
\setlength{\belowdisplayskip}{3pt}
\begin{eqnarray}
\mathcal{H}(\bm{\alpha ^i})=-\sum_{\alpha ^i_m\in \bm{\alpha ^i}}\mathcal{P}(\alpha ^i_m)\textrm{log}\mathcal{P}(\alpha ^i_m) .
\end{eqnarray}
Fig. 1 (b) shows the entropy variation curves of some layers in ResNet-20 on CIFAR-10 based on our single path search cell.
We can see that the entropies of different layers have different convergence speeds.
And many layers will gradually converge to a steady state in the searching process.
If we gradually determine the quantization precision of a certain layer in the searching process, the search space will decrease exponentially.
Therefore, we propose a sequential single path search method, which divides this task into subtasks by iterations and optimizes them sequentially.
After satisfying the decision conditions, we prioritize the cells with the highest selection certainty.
Then, we use the selected quantization precision to replace the original search cell to participate in the subsequent searching process.
A new search subproblem is generated by the above method.
With the determination of some search cells, the search space decreases exponentially.
Finally, a mixed-precision model satisfying the given constraints is obtained by iterative solutions.
The iterative procedure is shown in Algorithm 1.
In the searching process, the quantization precision of each search cell gradually tends to be stable and the entropy will gradually decrease through continuous iterative updating of architecture parameters.

\vspace{-3 mm}
\begin{algorithm}[!h]
    \setlength{\belowcaptionskip}{-0.2cm}
	\caption{: Sequential Single Path Search Method}
	\begin{algorithmic}[1]
		\REQUIRE 	\,\\
        The sub-training dataset and validation dataset; \\
		The search space $\mathbf{S}$, supernet and constraints $c_1$ and $c_2$;\\
		Initialize architecture parameters $\mathbf{A}=\{\alpha^i_{m} \}$, $\mathbf{B}=\{\beta^i_{n} \}$ and supernet weights $\mathbf{W}$. \\
		\ENSURE \,\\
        The architecture with high accuracy under given constraints. \\
		\WHILE{not terminated}
		\STATE Using Eqs. $(5)\sim(8)$ to select the forward subnetwork; \\
		\STATE Update weights $\mathbf{W}$ by using the sub-training dataset; \\
		\STATE Using Eqs. $(5)\sim(8)$ to select the forward subnetwork; \\
		\STATE Update architecture parameters $\mathbf{A}$ and $\mathbf{B}$ by using the validation dataset; \\
		\IF {decision epoch}
		\STATE Calculate the selection certainty $\mathcal{H}(\bm{\alpha ^i})$ of each search cell by using Eqs. $(12)\sim(13)$; \\
		\STATE Determine the quantization precision of selected cell by probability; \\
		\STATE Remove the selected cell from the search space; \\
		\ENDIF
		\ENDWHILE
	\end{algorithmic}
\end{algorithm}

\vspace{-1 mm}
\section{Experiments}
\vspace{-1 mm}

In this section, we search the mixed-precision quantization models to verify the effectiveness of our method on two image classification benchmarks (CIFAR-10
and ImageNet) and an object detection benchmark (COCO). We first describe the details of our experimental implementations.
Then the experimental results of our method are presented to compare with state-of-the-art methods.
%Finally, we analyze the convergence of our SSPS method.

\vspace{-1 mm}
\subsection{Implementation Details}
\vspace{-1 mm}

We implement our method using Pytorch \cite{paszke2017automatic}, in which we can easily implement and debug quantization functions and NAS algorithms.
We use a hardware-friendly $uniform$ quantization function as the quantizer, therefore, the inference process can be efficiently implemented by bitwise operations (e.g., xnor and bitcount) to achieve model compression, computational acceleration and resource saving.
We quantize the weights $\mathbf{W}^i$ linearly into $n$-bit, which can be formulated as follows:
\setlength{\abovedisplayskip}{3pt}
\setlength{\belowdisplayskip}{3pt}
\begin{eqnarray}
\mathcal{Q}_n(\mathbf{W}^i) = \textrm{round}(\textrm{clamp}(\mathbf{W}^i/t^i, 1)*(2^{n-1}-1))*d^i,
\end{eqnarray}
where the clamp function is used to truncate all values into the range of $[-1, 1]$, and $t^i$ is a learned parameter of the $i$-th search cell.
The scaling factor $d^i$ is defined as: $d^i = t^i/(2^{n-1}-1)$.
The search space of each search cell is $\mathbf{s}^i = \{2,3,4,5,6\}$.

In implementation, we combine the weight search cell and activation value search cell into one layer-level search cell.
Therefore, there are 25 candidates for each layer.
The architecture parameters are optimized by Adam, and the initial learning rate is $3 \times 10^{-3}$.
Network parameters $\mathbf{W}$ are updated by SGD, the initial learning rate is $1 \times 10^{-3}$, and the weight decay is set to $1 \times 10^{-4}$.
For ImageNet, the batch size for all the networks is set to $1024$.
The super-parameter $\lambda$ is set to $0.05$.
Following the methods \cite{zhuang2020training,Courbariaux2016Binarized,zhou2016dorefa},
we quantize the first convolutional layer and the last fully-connected layer to $8$-bit.
We apply the pre-trained full-precision model to initialize the supernet and then the warm-up strategy is adopted.
After searching the desired network, we fine-tune the mixed-precision quantization model to get the final parameters.
For COCO detection, we use the mixed-precision architecture obtained by the ImageNet classification task as the backbone.
Our network is fine-tuned by SGD for 50K iterations with the initial learning rate $1 \times 10^{-3}$ and the batch size of 16 for 8 V100 GPUs.
The learning rate is decayed by a factor of 10 at iterations 30K and 40K, respectively.

\vspace{-1 mm}
\subsection{Experimental Results}
\vspace{-1 mm}

\subsubsection{CIFAR-10}
\vspace{-1 mm}
We focus on searching the mixed-precision quantization model under the given average weight bit and average operation bit of ResNet-20 on the CIFAR-10 dataset.
This dataset has 50K training images and 10K testing images.
We divide the training images into a sub-training dataset (25K images) and a validation dataset (25K images).
The sub-training dataset is used to update the weights of supernet and then the validation dataset is used to update architecture parameters.
After searching process, we use the whole training images to fine-tune the selected model.
\vspace{-0.1cm}
\begin{table}[h]
	\centering
	\caption{Accuracy comparisons of ResNet20 on CIFAR10. Here 'M' refers to mixed-precision quantization models.}
    \vspace{-0.2cm}
	\setlength{\tabcolsep}{0.6mm}{
		\renewcommand\arraystretch{1.1}
		\begin{tabular}{cccccc}
			\hline
			Methods    & W-Bits    & A-Bits    & Top-1    & W-Comp   & Ave-Bits  \\
			\hline
			Baseline   & 32    & 32  & 92.37    & 1.00        &  32.00    \\
			\hline
			Dorefa \cite{zhou2016dorefa}    & 3     & 3   & 89.90    & 10.67       &  3.00    \\
			PACT  \cite{choi2018pact}     & 3     & 3   & 91.10    & 10.67       &  3.00     \\
			LQ-Nets \cite{zhang2018lq}   & 3     & 3   & 91.60    & 10.67       &  3.00    \\
			HAWQ  \cite{dong2019hawq}     & M     & 4   & 92.22    & 13.11       &  -     \\
			BP-NAS \cite{yu2020search}    & M     & M   & 92.12    & 10.74       &  3.30     \\
			\hline
			SSPS       & M     & M   & \textbf{92.54}    & 10.74       &  3.04     \\
			\hline
		\end{tabular}
	}
\vspace{-0.3cm}
\end{table}

\begin{table}[!t]
    \setlength{\abovecaptionskip}{0.4cm}
    \footnotesize
	\centering
	\caption{Accuracy comparisons of ResNet-18, ResNet-34, ResNet-50 and MobileNet-V2 on ImageNet. Here 'M' refers to mixed-precision
		quantization models.}
    \vspace{-0.2cm}
	\setlength{\tabcolsep}{0.6mm}{
		\renewcommand\arraystretch{1.2}
		\begin{tabular}{ccccccc}
			\hline
			Models & Methods    & W-Bits    & A-Bits    & Top-1    & W-Comp   & Ave-Bits  \\
			\hline
            \multirow{12}*{ResNet-18}
			&Baseline   & 32    & 32  & 70.20    & 1.00      & 32.00     \\ \cline{2-7}
			&PACT \cite{choi2018pact}      & 3     & 3   & 68.10    & 10.67     &   3.00   \\
			&LQ-Nets \cite{zhang2018lq}   & 3     & 3   & 68.20    & 10.67     &   3.00    \\
			&DSQ \cite{gong2019differentiable}       & 3     & 3   & 68.66    & 10.67     &  3.00     \\
			&QIL  \cite{jung2019learning}      & 3     & 3   & 69.20    & 10.67     &  3.00      \\ \cline{2-7}
			&SSPS       & M     & M   & \textbf{69.64}    & 10.65     &  2.99     \\ \cline{2-7}
			&PACT \cite{choi2018pact}      & 4     & 4   & 69.20    & 8.00      &  4.00   \\
			&LQ-Nets \cite{zhang2018lq}   & 4     & 4   & 69.30    & 8.00      & 4.00      \\
			&DSQ  \cite{gong2019differentiable}      & 4     & 4   & 69.56    & 8.00      &  4.00     \\
			&QIL  \cite{jung2019learning}      & 4     & 4   & 70.10    & 8.00      &  4.00     \\
			&AutoQ \cite{lou2019autoq}     & M     & M   & 68.20    & 6.91      &  -    \\ \cline{2-7}
			&SSPS       & M     & M   & \textbf{70.70}    &  7.95     &  3.95        \\
			\hline
            \multirow{10}*{ResNet-34}
			&Baseline   & 32    & 32  & 73.8     & 1.00      & 32.00     \\ \cline{2-7}
			&ABC-Net \cite{lin2017towards}   & 3     & 3   & 66.70    & 10.67     & 3.00 \\
			&LQ-Nets \cite{zhang2018lq}   & 3     & 3   & 71.90    & 10.67     & 3.00 \\
			&DSQ \cite{gong2019differentiable}       & 3     & 3   & 72.54    & 10.67     & 3.00 \\
            &QIL \cite{jung2019learning}       & 3     & 3   & 73.10    & 10.67     & 3.00 \\ \cline{2-7}
			&SSPS       & M     & M   & \textbf{73.49}    & 10.69     & 3.06      \\ \cline{2-7}
            &BCGD \cite{yin2019blended}      & 4     & 4   & 70.81    & 8.00      & 4.00 \\
            &DSQ  \cite{gong2019differentiable}      & 4     & 4   & 72.76    & 8.00      & 4.00 \\
            &QIL  \cite{jung2019learning}      & 4     & 4   & 73.70    & 8.00      & 4.00 \\ \cline{2-7}
			&SSPS       & M     & M   & \textbf{74.30}    & 7.99      & 4.01      \\
			\hline
            \multirow{6}*{ResNet-50}
			&Baseline   & 32    & 32  & 77.15    & 1.00      & 32.00     \\ \cline{2-7}
			&AutoQ \cite{lou2019autoq}     & M     & M   & 63.21    & 9.12      &  -    \\
			&HAQ  \cite{wang2019haq}      & M     & M   & 75.48    & -     & 3.60 \\
			&HAWQ \cite{dong2019hawq}      & M     & M   & 75.30    & -     & 4.00 \\
            &BP-NAS \cite{yu2020search}    & M     & M   & \textbf{76.67}    & -      & 3.80 \\ \cline{2-7}
			&SSPS       & M     & M   & 76.22    & 8.00      & 3.98      \\
			\hline
            \multirow{6}*{\!MobileNet-V2\!}
			&Baseline   & 32    & 32  & 71.87    & 1.00      & 32.00     \\\cline{2-7}
			&DSQ  \cite{gong2019differentiable}      & 4     & 4   & 64.80    & 8.00      & 4.00      \\
			&TQT  \cite{jain2019trained}      & 4     & 4   & 67.79    & 8.00      & 4.00      \\
			&HAQ  \cite{wang2019haq}      & M     & M   & 66.99    &  -        & -     \\
			&AutoQ \cite{lou2019autoq}     & M     & M   & 69.02    & 7.58      &  -    \\ \cline{2-7}
			&SSPS       & M     & M   & \textbf{69.10}    & 7.99      & 4.02    \\
			\hline
		\end{tabular}
	}
	\label{tab:cifar10table}
\vspace{-0.6cm}
\end{table}

%We compare the performance of our method with those of
%Dorefa \cite{zhou2016dorefa}, PACT \cite{choi2018pact}, LQ-Nets \cite{zhang2018lq}, HAWQ \cite{dong2019hawq} and BP-NAS \cite{yu2020search}.
For each compared method, we report its average weight bit, average activation value bit, Top-1 accuracy, model size compression rate and average operation bit.
The target average weight bit and average operation bit of searching process are $3.00$.
The results are shown in Table 1.
Compared with the full-precision model (Baseline),
our model outperforms it by up to 0.17\% while still achieving $10.74\times$ compression ratio for weights.
Compared with the $network$-$wise$ quantization methods, Dorefa, PACT and LQ-Nets, the Top-1 accuracy of our method increases by $2.64\%$, $1.44\%$ and $0.94\%$, respectively.
Similarly, our method has obvious advantages over the mixed-precision quantization methods.
Moreover, our method performs much better than HAWQ and BP-NAS, and its Top-1 accuracy increases by 0.32\% and 0.42\%, respectively.

\vspace{-3 mm}
\subsubsection{ImageNet}
\vspace{-1 mm}

In order to verify the search ability of our method on large-scale datasets and deep networks,
we implement ResNet-18, 34, 50 and MobileNet-V2 on the ImageNet (ILSVRC2012) dataset.
%This dataset consists of 1K categories images, and has over 1.2M images in the training dataset and 50K images in the validation dataset.
We choose three-quarters of the training dataset as the sub-training dataset to update the weights of the supernet.
The remaining one-quarter of the training dataset is used as the validation dataset to update the architecture parameters.

%We compare the performance of our method with those of state-of-the-art methods
%such as PACT \cite{choi2018pact}, LQ-Nets \cite{zhang2018lq}, DSQ \cite{gong2019differentiable}, QIL \cite{jung2019learning}, ABC-Net \cite{lin2017towards}, BCGD \cite{yin2019blended}, TQT \cite{jain2019trained}, HAQ \cite{wang2019haq}, HAWQ \cite{dong2019hawq}, BP-NAS \cite{yu2020search}, AutoQ \cite{lou2019autoq} and their baseline.
Table 2 shows the experimental results, where the method marked 'M' represents mixed-precision quantization.
Similar to other methods, our experiments mainly focus on the average 3-bit and 4-bit quantization.
In the searching process, we set the average weight bit and average operation bit to $3$ or $4$ to control the search direction.
%For ResNet-18 and ResNet-34, we retrain the mixed-precision model for 30 epochs with the batch size of 1024 on 4 GPUs after completing  the searching process.
From Table 2, we can see that our selected mixed-precision quantization models of ResNet-18 and ResNet-34 achieve the best accuracies,
which are higher than their full-precision counterparts.
For ResNet-50, we compare our method with several mixed-precision quantization methods.
HAQ and AutoQ  apply reinforcement learning to search for mixed-precision quantized architectures.
They spend more time on training and the results are still worse than ours.
HAWQ manually chooses the bit precision among the reduced search space, and its result is 0.92\% lower than ours.
BP-NAS uses small sampled datasets to complete the searching process and then transfer it to ResNet-50.
Our method is still comparable with BP-NAS, although its results are obtained after 150 epochs of fine-tune with label smooth.
As a lightweight network, MobileNet-V2 eliminates many redundant calculations.
Therefore, the model quantization of MobileNet-V2 will bring great precision loss.
Even so, compared with DSQ, TQT, HAQ and AutoQ, our method can converge well and outperforms them by up to 4.3\%, 1.31\%, 2.11\% and 0.08\%, respectively.

\vspace{-1 mm}
\subsubsection{COCO Detection}
\vspace{-1 mm}

We further explore the effectiveness of our mixed-precision model for detection tasks on the COCO benchmark \cite{lin2014microsoft},
which is one of the most popular large-scale benchmark datasets for object detection.
This dataset consists of images in 80 different categories.
We use the trainval35k split for training and minival split for validation.
Both one stage RetinaNet \cite{lin2017focal} detector and two stage Faster R-CNN \cite{ren2015faster} detector are applied to verify the effectiveness of
our selected mixed-precision model.
In other words, we use the selected mixed-precision ResNet-50 model in Section 5.2.2 as backbones.
For Faster R-CNN, the RPN and ROIhead are quantized to 4-bit.
For RetinaNet, the feature pyramid and detection heads are quantized to 4-bit, except for the last layer in the detection heads is quantized to 8-bit.

\begin{table}[h]
    \setlength{\abovecaptionskip}{0.2cm}
    \footnotesize
	\centering
	\caption{Results of Faster R-CNN and RetinaNet on the COCO validation dataset.}
    %\vspace{-0.2cm}
	\setlength{\tabcolsep}{0.7mm}{
		\renewcommand\arraystretch{1.2}
		\begin{tabular}{cccccccccc}
			\hline
            \multicolumn{9}{c}{ResNet-50 + Faster R-CNN} \\
			\hline
			Methods  & W/A-Bits & Ave-Bits & AP    & $\textrm{AP}_{50}$    & $\textrm{AP}_{75}$    & $\textrm{AP}_{S}$ & $\textrm{AP}_{M}$ & $\textrm{AP}_{L}$  \\
            \hline
			Baseline& 32/32 & 32.00  & 37.7  & 59.3   & 40.9  & 22.0  & 41.5   & 48.9        \\
            \hline
			FQN \cite{li2019fully}    & 4/4   & 4.00  & 33.1   & 54.0   & 35.5  & 18.2  & 36.2  & 43.6      \\
			BP-NAS \cite{yu2020search} & M/M   & $\sim$4.00  & 35.8   & 57.9   & 38.3   & 21.7  & 39.8  & 47.4      \\
            SSPS   & M/M    & $\sim$4.00  & \textbf{37.4}   & {58.1}   & {40.6}   & {22.1}  & {40.4}  & {47.9}       \\
            \hline
            \multicolumn{9}{c}{ResNet-50 + RetinaNet} \\
			\hline
			Methods  & W/A-Bits & Ave-Bits & AP    & $\textrm{AP}_{50}$    & $\textrm{AP}_{75}$    & $\textrm{AP}_{S}$ & $\textrm{AP}_{M}$ & $\textrm{AP}_{L}$  \\
            \hline
			Baseline & 32/32 & 32.00  & 37.8   & 58.0   & 40.8   & 23.8  & 41.6  & 48.9       \\
            \hline
			FQN \cite{li2019fully}    & 4/4   & 4.00  & 32.5   & 51.5   & 34.7   & 17.3  & 35.6  & 42.6      \\
			Auxi \cite{zhuang2020training}   & 4/4   & 4.00  & 36.1   & 55.8   & 38.9   & 21.2  & 39.9  & 46.3      \\
            SSPS   & M/M    & $\sim$ 4.00    & \textbf{36.4}  &  {55.8} & {38.6 } & {20.8}  &  {39.9}  &  {47.6}      \\
			\hline
		\end{tabular}
	}
	\label{tab:Resnettable}
    \vspace{-2 mm}
\end{table}

We compare the performance of our method with those of FQN \cite{li2019fully}, Auxi \cite{zhuang2020training} and BP-NAS \cite{yu2020search},
where FQN and Auxi are fixed-precision methods, and BP-NAS is a mixed-precision method.
Table 3 shows the experimental results.
As we can see, model quantization can affect the detection results, obviously.
For Faster R-CNN detector, our selected mixed-precision model demonstrate better performance.
For example, our 4-bit detector with ResNet-50 backbone outperforms FQN and BP-NAS by $4.3\%$ and $1.6\%$, respectively.
For RetinaNet detector, our model also shows the best experimental results in quantized models.

\vspace{-1 mm}
\subsection{Effective and Fast}
\vspace{-1 mm}
\subsubsection{Convergence Analysis}
\vspace{-1 mm}

We take the mixed-precision search of ResNet-18 as an example for convergence analysis,
where we set the expected average weight bit and average operation bit to 4.
Fig. 4 show the convergence curves of the searching loss, average operation bit and average weight bit.
\textbf{Just as an ablation experiment,
the blue line represents the convergence curve of our SSPS method,
and the orange line represents the convergence curve without the decision operations (step 6 $\sim$ 10 in Algorithm 1), which is called SPS.}
%SSPS takes less than 7 hours on 4 V100 GPUs and SPS takes more than 11 hours on 4 V100 GPUs.
From Fig. 4 (a), we can see that our decision operations can significantly improve the convergence of the searching process.
The decision operations can reduce the search space and the factors affecting the target, thus increasing the stability of the searching process.
%At the beginning, the temperature coefficient $\tau$ is relatively large, which leads to the fluctuation of the target values.
With the decrease of temperature coefficient $\tau$ and the convergence of training loss, the fluctuation becomes smaller and smaller until they approach to the target constraints.
Fig. 5 shows the selected quantization policy for ResNet-18.
The histogram represents the quantization precision of each layer in our model.
The upper part represents the weight precision of different layers,
and the bottom part represents the quantization precision of activation values in different layers.
In particular, we specially mark the order of precision decision for the layers in the process of our model search.

\begin{figure}[!htb]
    \setlength{\abovecaptionskip}{0.1cm}
    \setlength{\belowcaptionskip}{-0.4cm}
    \centering
	\includegraphics[width=3.1in]{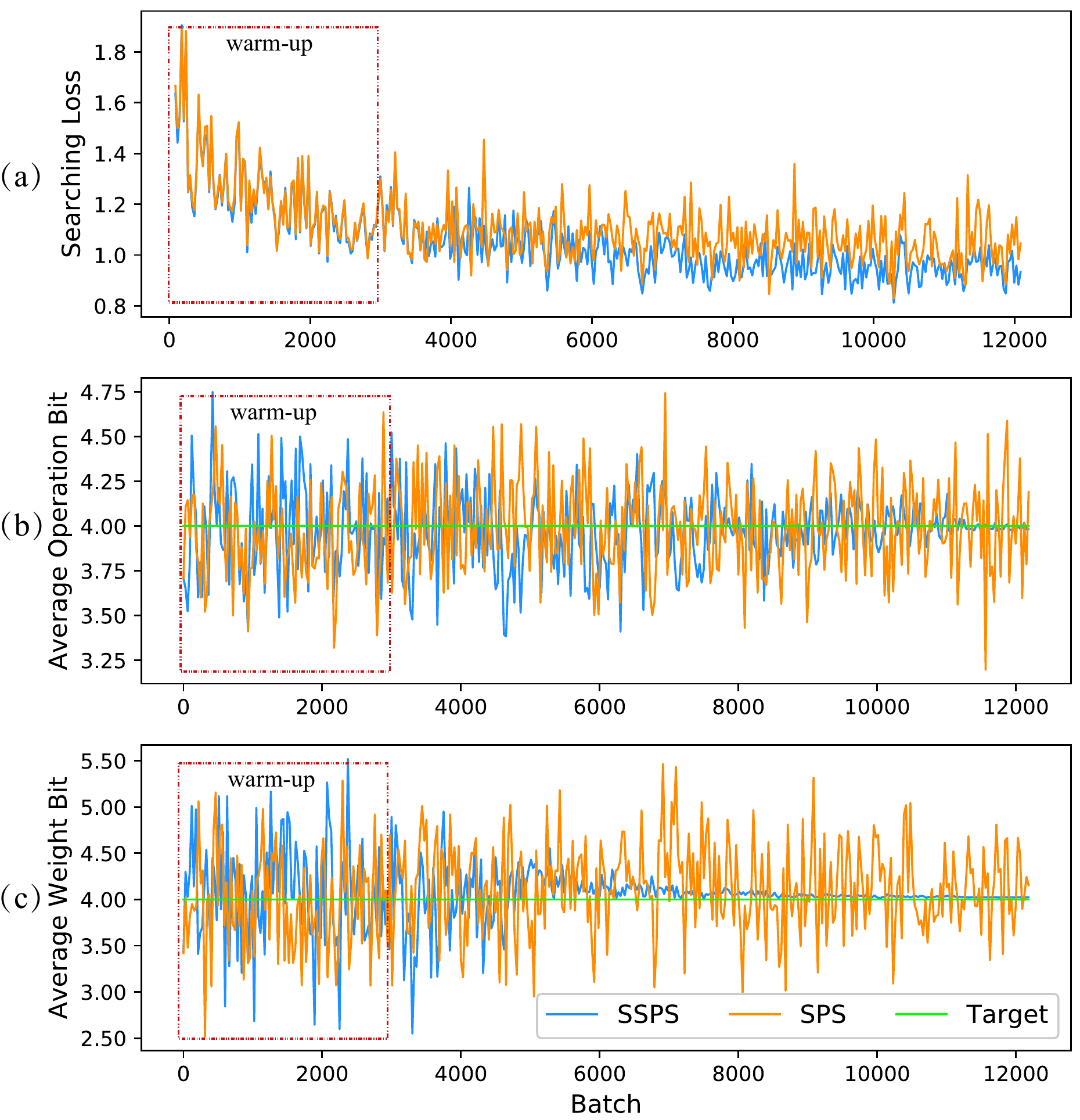}
	\caption{\small{(a) The convergence curve of searching loss.
    (b) The convergence curve of average operation bit.
    (c) The convergence curve of average weight bit.}
    }
\end{figure}

\begin{figure}[!htb]
    \setlength{\abovecaptionskip}{0.1cm}
    \setlength{\belowcaptionskip}{-0.4cm}
    \centering
	\includegraphics[width=3.1in]{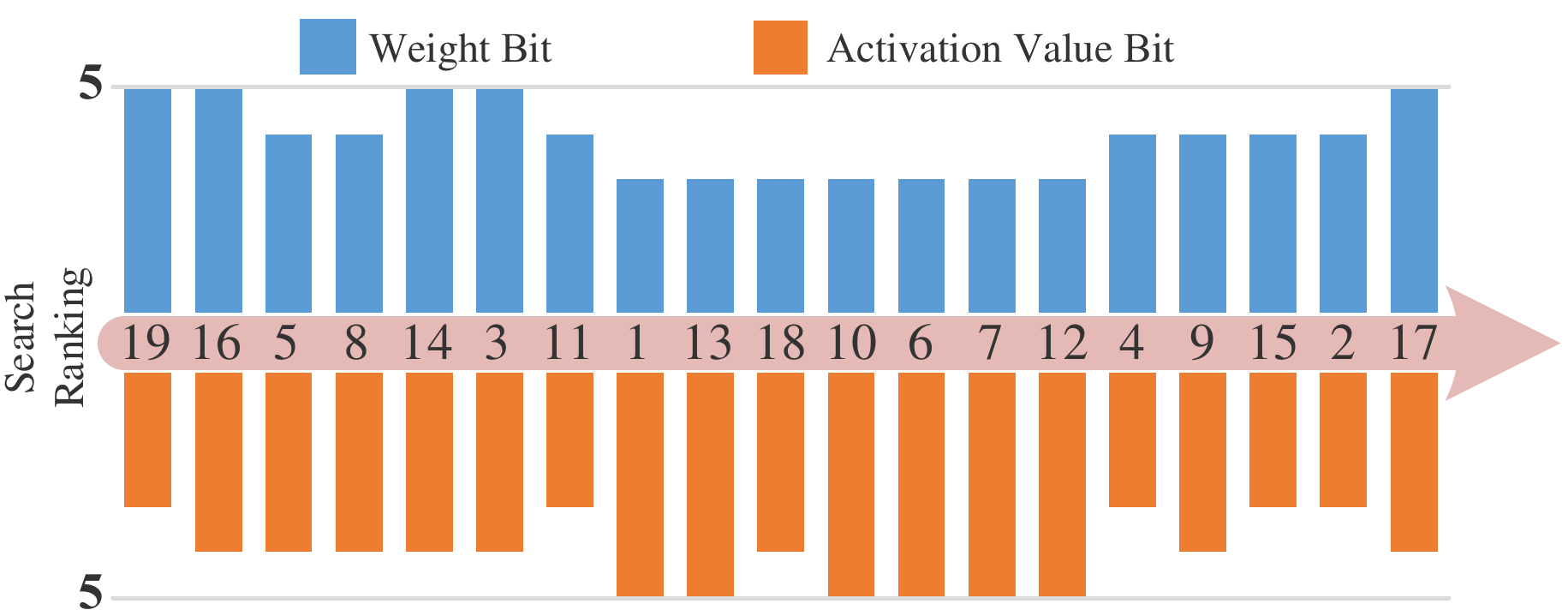}
	\caption{\small{Selected quantization policy for ResNet-18.
    The Y-axis represents the quantization precision, and the X-axis represents the order of precision decision for the layers.
    The numbers in the arrow indicate the order in which the layers are determined.}
    }
\end{figure}

\vspace{-2 mm}
\subsubsection{Comparison with Related Work}
\vspace{-2 mm}
In this subsection, we discuss the differences of our mthod with two similar methods, DNAS \cite{wu2018mixed} and BP-NAS \cite{yu2020search}.

\textbf{Compared with DNAS:}
(1) By using the Gumbel Softmax with an annealing temperature, the pipeline behaves of DNAS is very similar to DARTS at the beginning,
in which multiple candidates participate in the calculation.
Therefore, it takes up a lot of memory resources just like DARTS, as shown in Fig. 1 (a).
However, only one candidate is allowed to pass through our search cell, thus saving hardware resources.
(2) DNAS introduces the cost of candidate structures into the loss function to encourage using lower-precision weights and activations.
Because there is no target setting, DNAS needs multiple searches to get an appropriate model.
Note that, our method only needs once search to return an optimal model under given constraints.
(3) Our sequential search method can exponentially reduce the search space in the search process, thus improving the search speed and convergence stability, as shown in Fig. 4.
 Our method takes less than 28 GPU-hours to complete a search for ResNet18 on ImageNet, which is much faster than DNAS (40 GPU-hours).
(4) DNAS is a block-wise mixed-precision method whose all layers in one block use the same precision.
Our method is a layer-wise mixed-precision method, which allows different levels of precision in the same block.

\textbf{Compared with BP-NAS:}
(1) BP-NAS applies DARTS to address optimization problems, which leads to a sharp increase in resources demand.
(2) For ImageNet, BP-NAS needs to randomly sample 10 categories and takes 5000 images as the training dataset to search.
This method relies on the incomplete definition of mixed-precision quantization tasks, which are easy to fall into a local optimal solution.
(3) Similar to DNAS, BP-NAS is also a block-wise mixed-precision method. The search space is much smaller than our method.

\vspace{-2 mm}
\section{Concluding Remarks}
\vspace{-2 mm}
In this paper, we proposed a novel SSPS method for mixed-precision quantization search and introduced constraints into our search loss function to guide the searching process.
This is a fully differentiable model, and the searching process can be optimized by gradient descent methods.
In the searching process, we determined the quantization precision according to the selection certainty of search cells,
which can reduce the search space exponentially and accelerate the search convergence speed.
Experimental results demonstrated that our proposed SSPS method achieves better
testing performance with similar constraints, compared to state-of-the-art methods on CIFAR-10, ImageNet and COCO.
Our future work will focus on mixed-precision quantization architecture search without training datasets and training a universal model that can support multiple quantization precision
to meet more industrial demands.

{\small
\bibliographystyle{ieee_fullname}
\bibliography{egbib}
}

\end{document}